\pdfoutput=1

\documentclass[11pt]{article}

\usepackage[final]{acl}

\usepackage{times}
\usepackage{latexsym}
\usepackage[normalem]{ulem}
\usepackage{tabularx}
\usepackage{multirow}
\usepackage[most]{tcolorbox}
\usepackage{booktabs}
\usepackage{adjustbox}
\usepackage{cleveref}

\usepackage[T1]{fontenc}

\usepackage[utf8]{inputenc}

\usepackage{microtype}

\usepackage{inconsolata}

\usepackage{graphicx}

\title{Semantic Captioning: Benchmark Dataset and Graph-Aware Few-Shot In-Context Learning for SQL2Text}

\author{Ali Al-Lawati ~~~~~~~~~~ Jason Lucas ~~~~~~~~~~ Prasenjit Mitra \\
  The Pennsylvania State University \\ University Park, PA, USA \\
  \small{\{\texttt{aha112,jsl5710,pmitra\}@psu.edu}}
}

\begin{document}
\maketitle

\begin{abstract}

Large Language Models (LLMs) have demonstrated remarkable performance in various NLP tasks, including semantic parsing, which translates natural language into formal code representations. However, the reverse process, translating code into natural language, termed \textit{semantic captioning}, has received less attention. This task is becoming increasingly important as LLMs are integrated into platforms for code generation, security analysis, and educational purposes. In this paper, we focus on the captioning of SQL query (SQL2Text) to address the critical need for understanding and explaining SQL queries in an era where LLM-generated code poses potential security risks. We repurpose Text2SQL datasets for SQL2Text by introducing an iterative ICL prompt using GPT-4o to generate multiple additional utterances, which enhances the robustness of the datasets for the reverse task. We conduct our experiments using in-context learning (ICL) based on different sample selection methods, emphasizing smaller, more computationally efficient LLMs. Our findings demonstrate that leveraging the inherent graph properties of SQL for ICL sample selection significantly outperforms random selection by up to 39\% on BLEU score and provides better results than alternative methods. 
Dataset and codes are published online\footnote{Code/Dataset Repository: \url{https://github.com/aliwister/ast-icl}}.
\end{abstract}

\section{Introduction} 

Large Language Models (LLMs) have advanced significantly, exhibiting paradigm-shifting capabilities in natural language processing (NLP). Notably, in-context learning (ICL) \cite{zhao2023survey} and semantic reasoning \cite{tang2023large} abilities allow LLMs to excel in natural language understanding (NLU) and generation tasks without extensive supervised learning.
This has enabled LLMs, such as ChatGPT~\cite{openai2024chatgpt}, to demonstrate remarkable capabilities in generating correct code without additional training or fine-tuning. As a result, LLMs have been widely adopted in the programming community as code explainers and commenters~\cite{chen2023gptutor}. However, there is a notable gap in the literature regarding how well these LLMs understand code.

\begin{figure}
    \centering
    \includegraphics[width=1.0 \linewidth]{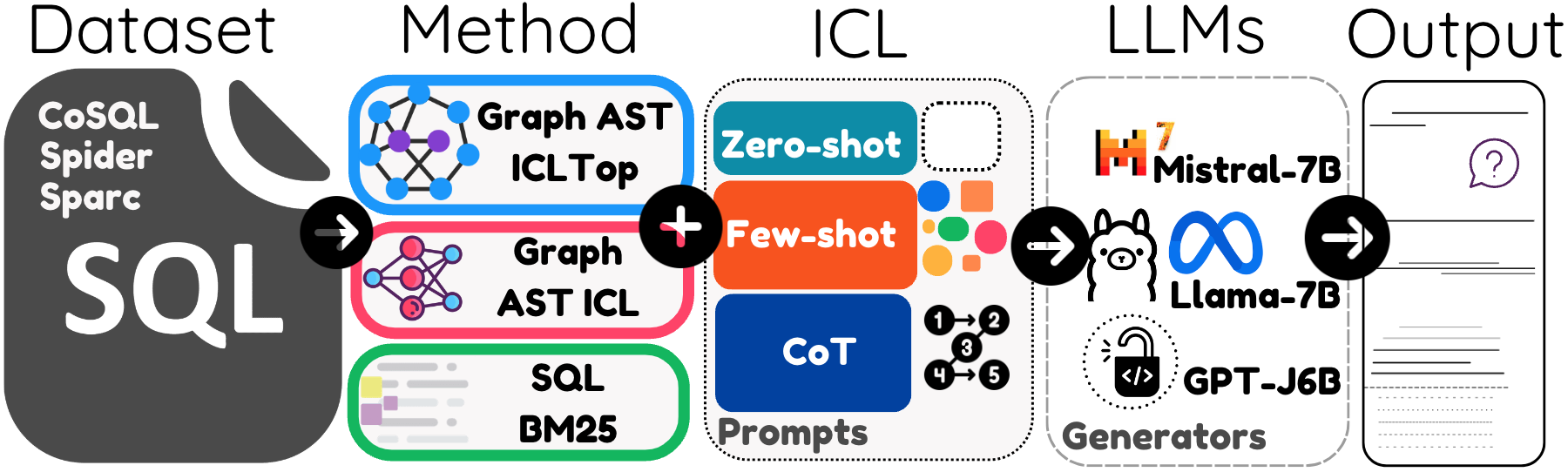}
    \caption{Graph AST-ICL Framework for SQL2Text generation showing the pipeline from SQL input through Graph-AST methods and ICL approaches to final text output using various LLM generators (e.g., Mistral).
    }
    \label{fig:flow}
   \vspace{-19pt}
\end{figure}

Semantic parsing (Text2SQL), a popular downstream task has been extensively used to assess the capacity of language models to generate correct formal code segments from corresponding natural language text, e.g., using business-related questions to generate Structured Query Language (SQL) codes~\cite{berant2014semantic, li2024can}.    
However, there is a gap in the evaluation methods and datasets for assessing generated codes. In particular, SQL2Text, which can be seen as a specific form of code summarization or translation. While Text2SQL is well-developed and studied using LLMs~\cite{li2024can, zhang2024sqlfuse}, SQL2Text remains underexplored, particularly in the ICL setting.  This is due to the diversity of code domains and the limited availability of suitable datasets. 
This leads to a natural question: \uline{if LLMs are good at translating natural language to code, how well do they perform on the reverse task}?  

Therefore, we consider the domain of formal code and introduce a new task called \textit{semantic captioning} as the reverse operation of semantic parsing. 
Semantic captioning can be defined as an SQL2Text operation, where, given an SQL query, an LLM translates the formal code into a corresponding natural language question (i.e., utterance or caption). We focus on SQL code understanding because SQL is the most commonly used language for data extraction, ranking \# 1 on IEEE Spectrum's 2023 ``Jobs List'' \cite{zhang2024sqlfuse}. Also, SQL was recently reported among the 8 most popular programming languages in 2024~\cite{tiobe2024index}, underscoring its widespread importance across various technology and business roles. However, the literature mainly focuses on advancing NLU in popular high-resource programming languages, such as Python~\cite{nam2024using}. 

As LLM adoption grows, bidirectional SQL-LLM translation (SQL2Text and Text2SQL) becomes crucial. This capability is vital for system security, software development, and education, such as building automated support tools for risk assessment and assistive learning. Therefore, as businesses increasingly integrate AI-powered applications and automate tasks using LLMs, the ability to generate accurate and meaningful descriptions of SQL queries becomes increasingly relevant. 

Thus, to investigate these gaps in knowledge, we formulate two research questions (RQs) to study semantic captioning: 

\noindent \textbf{RQ1: Can we leverage existing Text2SQL dataset for semantic captioning?} We develop the first semantic captioning dataset using iterative ICL prompt-based techniques.

\noindent \textbf{RQ2: Can leveraging SQL's graph properties for sample selection improve ICL performance?} We propose a graph-aware method and compare it to zero-shot and few-shot approaches, and we investigate Smaller LLM (e.g., Mistral) vs. Larger LLM (GPT-4) performances across all methods.

First, we use GPT-4o and vetted Text2SQL corpora to generate the first dataset for semantic captioning using a novel iterative ICL prompt on three benchmark datasets: CoSQL \cite{yu2019cosql}, Spider~\cite{yu2018spider}, and SParC~\cite{yu2020sparc}. Next, we perform a human expert evaluation on the generated and original datasets to maintain quality. Finally, we experiment with three smaller-sized LLMs: GPT-J-6B~\cite{gpt-j}, Mistral-7B~\cite{jiang2023mistral}, and CodeLlama-7B \cite{roziere2023code}; and one larger LLM: GPT-4~\cite{achiam2023gpt}. In the ICL setting, we leverage the graph properties of SQL using Graph Neural Network (GNN) and graph pooling of the SQL abstract syntax tree (AST) to select relevant prompt demonstrations. To assess our models' performance, we utilize the vetted datasets using four evaluation metrics: BLEU-4~\cite{papineni2002bleu}, two variations of BERTScore~\cite{zhang2019bertscore}, and AlignScore~\cite{zha2023alignscore} to assess the quality of LLM-generated semantic captions.

Our contributions include: (1) introducing semantic captioning, a new task focusing on SQL code understanding; (2) contributing three benchmark datasets that have been repurposed from existing Text2SQL benchmarks (Spider, CoSQL, and SparC), processed using an iterative generation for additional utterances, and evaluated using human experts; (3) proposing a novel graph-based ICL method that enhances the performance of smaller-sized LLMs; and (4) conducting experiments comparing zero-shot, few-shot (random and BM25) with our method, as well as comparing smaller vs. larger LLMs. 
Our findings advance the understanding of LLMs' ability to comprehend SQL queries and generate corresponding natural language questions, facilitating the integration of LLMs into platforms, with implications for improving code security, automating documentation, and enhancing educational tools. 

\begin{table*}[h!]
   \centering
   \begin{adjustbox}{width=1\textwidth,center}
   \begin{tabular}{l|l|ccrr|ccrr|ccrr}
   \toprule
   \multirow{2}{*}{Language Model} & 
   \multirow{2}{*}{Method} & 
   \multicolumn{4}{c|}{CoSQL-S2T} &  
   \multicolumn{4}{c|}{SparC-S2T} & 
   \multicolumn{4}{c}{Spider-S2T}   \\ 
   & &
    BERTScore-1 & BERTScore-2 & BLEU & AlignScore & 
    BERTScore-1 & BERTScore-2 & BLEU & AlignScore &
    BERTScore-1 & BERTScore-2 & BLEU & AlignScore \\
   \midrule
   \multirow{6}{*}{GPT-J}  
        & Zero-shot 
& 0.5831 & 0.4894 & 0.0061 & 0.4585
& 0.6158 & 0.5009 & 0.0084 & 0.4885
& 0.6177 & 0.4970 & 0.0096 & 0.5419\\

        & Random-2 
& 0.7837 & 0.7392 & 0.2021 & 0.4943
& 0.8128 & 0.7619 & 0.2472 & 0.5346
& 0.8207 & 0.7778 & 0.2561 & 0.6036 \\
        & BM25-2 
& 0.7779 & 0.7421 & 0.2197 & 0.5189
& 0.833$^\dagger$& 0.7845$^\dagger$ & 0.2890 & 0.5954
& 0.8271 & 0.7916 & 0.2799 & 0.6428 \\
        & \textbf{AST-ICL-TOP-2} 
& \textbf{0.8153}$^{\dagger\ddagger}$ & \textbf{0.7783}$^{\dagger\ddagger}$ & \textbf{0.2813}$^{\dagger\ddagger}$ & \textbf{0.6357}$^{\dagger\ddagger}$
& \textbf{0.8484}$^\dagger$ & \textbf{0.8151}$^{\dagger\ddagger}$ & \textbf{0.3093}$^\dagger$ & \textbf{0.6744}$^{\dagger\ddagger}$
& \textbf{0.8552}$^{\dagger\ddagger}$ & \textbf{0.8261}$^{\dagger\ddagger}$ & \textbf{0.3301}$^\dagger$ & \textbf{0.7201}$^{\dagger\ddagger}$
\\
        & \textbf{AST-ICL-2} 
& 0.8082$^{\dagger\ddagger}$ & 0.7711$^{\dagger\ddagger}$ & 0.2376 & 0.5583
& 0.8037$^\ddagger$ & 0.7635 & 0.2294$^\ddagger$ & 0.5254$^\ddagger$
& 0.8312 & 0.7856 & 0.2741 & 0.6327
\\
   \midrule
   \multirow{6}{*}{Mistral}
        & Zero-shot 
& 0.7442 & 0.6669 & 0.2458 & 0.7464
& 0.7822 & 0.6878 & 0.2391 & 0.7787
& 0.7597 & 0.6717 & 0.2226 & 0.7717
        
\\
        & Random-2 
& 0.8515 & 0.8145 & 0.3509 & 0.7335
& 0.8751 & 0.8351 & 0.3760 & 0.7640
& 0.8750 & 0.8436 & 0.4138 & 0.8066
        
\\
        & BM25-2 
& 0.8498 & 0.8143 & 0.3415 & 0.7499
& 0.8797 & 0.8423 & 0.3928 & 0.8219$^\dagger$
& 0.8749 & 0.8442 & 0.4027 & 0.8275
        
 \\
        & \textbf{AST-ICL-TOP-2} 
& \textbf{0.8632} & \textbf{0.8306}$^{\dagger\ddagger}$ & \textbf{0.3732} & \textbf{0.8103}$^{\dagger\ddagger}$
& \textbf{0.8883} & \textbf{0.8572}$^\dagger$ & \textbf{0.3977} & \textbf{0.8259}$^\dagger$
& \textbf{0.8920}$^{\dagger\ddagger}$ & \textbf{0.8649}$^{\dagger\ddagger}$ & \textbf{0.4348} & \textbf{0.8645}$^\dagger$
        
 \\
        & \textbf{AST-ICL-2} 
& 0.8577 & 0.8200 & 0.3369 & 0.7382
& 0.8707 & 0.8386 & 0.3727 & 0.7957
& 0.8801 & 0.8462 & 0.3861 & 0.8120
        
 \\
   \midrule
   \multirow{6}{*}{CodeLlama} 
        & Zero-shot 
& 0.7285 & 0.6740 & 0.1704 & 0.6718
& 0.7623 & 0.6979 & 0.1871 & 0.6766
& 0.7470 & 0.6810 & 0.1716 & 0.6985    
 \\
        & Random-2 
& 0.8384 & 0.8019 & 0.2996 & 0.6756
& 0.8644 & 0.8269 & 0.3535 & 0.7067
& 0.8618 & 0.8318 & 0.353 & 0.7763\\
        & BM25-2 
& 0.8350 & 0.8022 & 0.3314 & 0.7153
& 0.8715 & 0.8320 & 0.3841 & 0.7585
& 0.8732 & 0.8471 & \textbf{0.3769} & 0.7709
 \\
        & \textbf{AST-ICL-TOP-2} 
& \textbf{0.8571}$^{\dagger\ddagger}$ & \textbf{0.8217}$^{\dagger\ddagger}$ & \textbf{0.3578}$^\dagger$ & \textbf{0.7766}$^{\dagger\ddagger}$
& \textbf{0.8818}$^\dagger$ & \textbf{0.8502}$^{\dagger\ddagger}$ & \textbf{0.3945} & \textbf{0.799}$^\dagger$
& \textbf{0.8780}$^\dagger$ & \textbf{0.8509}$^\dagger$ & 0.3725 & \textbf{0.8396}$^{\dagger\ddagger}$
 \\
        & \textbf{AST-ICL-2} 
& 0.8451 & 0.8085 & 0.3218 & 0.7031
& 0.8692 & 0.8348 & 0.3601 & 0.7567
& 0.8733 & 0.8413 & 0.3615 & 0.8126
 \\
\bottomrule
   \end{tabular}
   \end{adjustbox}
      \caption{Test Results for three benchmark datasets on three LLMs using different sample selection methods for ICL. The number after the hyphen refers to the number of demonstration samples used in the prompt, which is two in this experiment. $^\dagger$ denotes t-test statistical significance compared to Random. $^\ddagger$ denotes t-test statistical significance compared to BM25.}
      \label{tab:rq11}
\end{table*}

\section{Related Work}
\textbf{Prompt Selection} Semantic Parsing, including Text2SQL, is a popular downstream task for LLMs~\cite{zhang2020m}, which has been extensively evaluated using zero-shot, multi-shot, and based on other approaches such as fine-tuning~\cite{chang2024sv2}, and instruction-tuning~\cite{sun2023instruction}. Recently, ICL has shown the capacity to train LLMs using prompts by providing a few demonstration examples~\cite{brown2020language}. However, LLMs are sensitive to demonstration examples, which has led researchers to investigate new demonstration retrieval methods such as word overlap~\cite{askari2023injecting}. 

Several works~\cite{min_rethinking_2022, xie_explanation_2022, saunshi_mathematical_2021}  attempt to understand why ICL works and gauge its effectiveness to motivate the selection of best fitting demonstrations~\cite{li_unified_2023, luo_dricl_2023}.~\citet{qin_-context_2023} utilize the LLM to compare the reasoning path of training demonstrations with test examples. \citet{khalifa_exploring_2023} propose an ensemble approach that combines multiple methods and averages the predictions. Other techniques have relied on the similarity between the test sample and the demonstrations using simple text or embedding-based comparisons~\cite{luo_dricl_2023}. Recently,~\citet{chen_self-icl_2023} have suggested asking the LLM to provide tailored demonstrations for a given query. However, this inefficient approach may result in higher costs as it relies on a powerful LLM to suggest effective samples.    

On the contrary, there is very limited research on the reverse task: semantic captioning. A few researchers have modeled the SQL2Text problem as a seq-2-seq model~\cite{camara2024large}. ~\citet{xu2018sql} propose a seq-2-seq model that processes the SQL using GNNs. This improved the task compared to a simpler joint training of a SQL/Text pair. However, there is a gap in the literature on a comprehensive evaluation of LLMs on this task.

\textbf{ICL Commentation}
LLMs are increasingly used for code generation, and there has been a growing interest in their capacity to understand code~\cite{nam_using_2024}.~\citet{geng_large_2024} study how well LLMs can provide comments based on the number of demonstrations, whereas we consider how to select the most effective demonstrations. On the other hand, \citet{gao_what_2023} perform an experimental study on the effect of selection order, and the number of demonstrations on the outcome for different code tasks. 

\textbf{Iterative Prompt Engineering} Recent LLM advancements have introduced new ICL capabilities, shifting generative prompt-based approaches from one-step to multi-step (iterative) pipelines. This transition addresses LLMs' generative quality issues like hallucinations and overlooked details \citep{adams2023sparse, zhang2023summit}. Notable innovations include the Chain of Density (CoD) method by \citet{adams2023sparse}, producing detailed, entity-centric summaries with improved abstractiveness and reduced bias; an iterative single-agent approach for summary generation and refinement by \citet{zhang2023summit}; and a tri-agent LLM-based method enhancing customization by \citet{xiao2023chatgpt}. However, these techniques have not been applied to improve Text2SQL and SQL2Text generation. Our work fills this gap by leveraging an iterative generation technique to produce high-quality semantic captions for SQL queries.

\section{Problem Definition} In this section, we describe and define our iterative prompt-based generation task, which leverages LLMs' advanced capabilities, and our graph-aware semantic captioning, which improves LLMs' generative capabilities. \Cref{fig:graph} shows an overview of our methodological framework.

\subsection{Iterative Generation} First, we define our iterative prompt-based generative task that uses ICL and iterative generation to produce high-quality semantic captions. To gauge the performance of the SQL2Text problem, it is imperative to have datasets that support this problem. Existing Text2SQL datasets fail to cater well to this problem since a SQL query may be described in several different ways using natural language. In order to repurpose these datasets, we investigate an iterative process-based approach to generate and refine multiple utterances for a given query. Next, we verify that this approach consistently affects the test results of ICL. It is important to note that simple paraphrasing and one-step ICL generation such as SumCoT~\cite{wang2023element} are prone to inconsistencies.

\noindent \textbf{Definition 1.} We define our iterative prompt-based ICL approach for generating high-quality semantic captions as a three-step process. Let $x$ be an input SQL query. Our approach consists of:

\begin{itemize} 

\item Generation: Produce a set $\mathbf{U} = \{u_1, u_2, u_3\}$ of three distinct utterances or questions accurately representing the input SQL query $x$. 

\item Feedback: Generate a feedback set $\mathbf{F} = \{f_1, f_2, f_3\}$ corresponding to each utterance in $ \mathbf{U}$. Each $f_i$ identifies errors, ambiguities, or areas for improvement based on four criteria: accuracy, clarity, fluency, and relevance to the original SQL query $x$. 

\item Refinement: Produce a refined set of utterances $\mathbf{U'} = \{u'_1, u'_2, u'_3\}$ by applying the feedback in $\mathbf{F}$ to the original set $\mathbf{U}$, optimizing for accuracy, clarity, fluency, and relevance. \end{itemize}

This iterative process aims to maximize the quality of the semantic captions while ensuring an LLM can accurately use them for the SQL2Text generation task. Given an inference LLM, $G$, the refined utterances $u'_i \in \mathbf{U'}$ should reconstruct the original SQL query $x$ when used as input to $G$. $G(u'_i)$ should yield $x'$, which is semantically equivalent to $x$.

\subsection{Semantic Captioning} 

Second, we define our proposed method, leveraging graph properties for few-shot sampling to improve LLMs' semantic captioning capabilities using SQL2Text. We experiment with multiple selection techniques to provide ICL with a better set of demonstration samples. 
In particular, SQL can be naturally represented as an AST: a graph-based knowledge structure. First, we assess whether a graph embedding method can best guide the sample selection problem and enhance performance compared to random or word similarity selection approaches such as BM25~\cite{robertson2009probabilistic}. Then, we benchmark the performance improvements against zero-shot. Next, we measure how well our GNN method helps small models compared to large models such as GPT-4. Finally, we verify whether any improvement is garnered by applying our method to a large model. 

\noindent \textbf{Definition 2.} We define semantic captioning with prompt-based few-shot ICL for text generation as follows. Given a demonstration dataset \( \mathcal{M} \) containing pairs of SQL queries and corresponding natural language question examples $(x, y)$, we aim to construct an effective prefix input prompt $(\mathcal{P})$. Our prefix prompt template comprises three parameters $\mathcal{P}(t + e + s)$: instructional text $(t)$, ICL SQL examples $(e)$, and an SQL seed $(s)$. 

We utilize a knowledge-aware representation retriever \( R \) to select appropriate SQL query ICL example pairs \( \mathbf{N} = \{(x_1, y_1), \ldots, (x_n,y_n)\} \) from the dataset \( \mathcal{M} \) to generate a text sequence \( y' \) in the form of a question or utterance such that the similarity metric \( \mathcal{D}(y, y') \) is maximized. 

Given an inference LLM, \( G \), a good input prompt should lead to the target output sequence where the test example \( x_{\text{test}} \) is the SQL seed, $s$, concatenated to the prompt \( \mathcal{P} \) and passed as a prefix prompt to \( G \). Specifically, the decoding from the LLM \( G(\mathcal{P}; x_{\text{test}}) \) should yield \( y_{\text{test}} \).

\subsection{LLMs}

We employ LLMs for dataset generation and semantic captioning evaluation. For dataset generation, we employ OpenAI's GPT-4o\footnote{https://openai.com/index/hello-gpt-4o/} to create utterances from SQL queries in Text2SQL corpora. This LLM was selected for its proficiency in programming and natural language tasks.

For evaluation, we use: (1) smaller open-source LLMs: GPT-J-6B, Mistral-7B, and CodeLlama-7B; and (2) GPT-4 as a larger LLM for comparative evaluation. All models except GPT-4 are accessible via Hugging Face. Models details are provided in \Cref{appen:LLMs}.



\subsection{Datasets}
There is no existing dataset for this task, therefore we explore benchmark Text2SQL datasets. First, we verify that these Text2SQL datasets are applicable to our task. We vetted over 8 benchmark Text2SQL datasets. We verify their appropriateness by ensuring that each dataset has SQL and utterance question pairs and that each question/utterance pair does not contain schema elements such as tables or columns to avoid inconsistency, which can lead to leakage, providing additional context of the query schema that may make it easier for the language model to resolve. The author manually checked each sample across these datasets for inconsistencies. Finally, we selected the columns containing the text pairs (SQL and associated utterance), to repurpose them for the SQL2Text task. We selected three benchmarks Text2SQL datasets: CoSQL, Spider, and SParC as shown in \Cref{table:samples}, for our proposed task. 


\section{RQ1: Iterative ICL Generation}

To create the semantic captioning dataset, we processed the test sets of the selected datasets. For CoSQL, we leveraged the entire test set of 293, while for the other datasets, we randomly sampled 293 instances, totaling 879 samples for our generation task. We utilized OpenAI's recently released GPT-4o model for utterance generation. Through prompt engineering, we created a prompt template with a JSON structure that guides the LLM through three iterative processes: generate, review, and refine. Our iterative approach is similar to the CoD iterative generation method but employs feedback based on predefined criteria—\textit{accuracy}, \textit{clarity}, \textit{fluency}, and \textit{relevance}—following the dimensions defined by \citet{peyrard2019simple}. This feedback helps refine the initial generation using fewer steps. Additionally, our predefined JSON structure enhances generative controllability. The prompt consists of instructions to guide the LLM through the generation task (see \Cref{fig:CoC_prompt}). 

\subsection{Evaluation Metrics}
To validate the quality of our generated utterances, we assess both semantic and logical consistency to detect any hallucinations and semantic discrepancies. Following the F3 method proposed by \citet{lucas2023fighting}, we utilize (1) AlignScore for logical alignment and (2) BERTScore for semantic alignment.

AlignScore (0-1) assesses logical consistency between the original utterance of the SQL query and the generated one, with higher scores indicating better alignment. BERTScore (0-1) evaluates semantic similarity to a reference, with higher scores reflecting greater accuracy. Low scores in either metric suggest potential inconsistencies or inaccuracies. Details of these metrics are provided in~\Cref{appen:eval metrics}.


    

\subsection{RQ1 Results}
We generated a dataset of 2637 utterances; each of the 879 original SQL queries has a group of three utterances. We conducted an evaluation process comprising both automated evaluation and human assessment to ensure robust quality validation of semantic captioning.

\subsubsection{Generation Quality Check}

\begin{table}[t]
\centering

\label{tab:dataset_stats}
\begin{adjustbox}{width=0.5\textwidth,center}
\begin{tabular*}{.8\textwidth}{@{\extracolsep{\fill}}lccc}

\toprule
\textbf{Dataset} & \textbf{SQL Query} & \textbf{Original Utterance} & \textbf{Generated Utterance}\\
\midrule

CoSQL-S2T & 290 & 281 & 870 \\
SParC-S2T & 289 & 282 & 867\\
Spider-S2T & 274 & 270 & 822\\

\bottomrule
\end{tabular*}
\end{adjustbox}
\caption{Dataset Statistics post quality check. The new datasets were renamed with a `-S2T' postfix for clarity which abbreviates SQL2Text.}
\label{tab:final_dataset}
\end{table}

\paragraph{Automatic Evaluation}
For each dataset, we employed an iterative prompt-based approach that generates six references per SQL query: three initial and three final references. We compare the original human written utterance with the generated initial utterance (iterative prompt, step 1)  and refined final utterance (iterative prompt, step 3) using AlignScore and BERTScore. Our iterative prompt-based approach improved our average final AlignScore by 7\%. Similarly, we measure semantic consistency, showing BERTScore greater than 0.85 on average. We selected the top three references based on Alignscore and BERTScore to ensure our generated utterances maintain logical and semantic consistency with gold-label references. 

\paragraph{Human Expert Evaluation }
To ensure a more reliable technical assessment of the generated data, we prioritized domain expertise over crowd-sourcing by engaging three evaluators: two authors and one external expert with extensive NLP and database experience. The evaluators independently assessed a total of 879 queries across four datasets using a structured evaluation framework with the following criteria: (1) \textit{Semantic Accuracy:} Evaluating the correctness of natural language interpretation (2) \textit{SQL Logic Preservation:} Assessing retention of query constraints and operations and (3) \textit{Technical Precision:} Assessing accurate representation of SQL operations.

For each sample, we conducted a quality assessment using the SQL query, gold-label reference, and our selected top three generated references. The evaluation followed a four-point scale: (1) \textit{Full Inconsistency:} Generated samples are considered inconsistent when any references incorrectly reflect (semantically, logically, and technically) all key elements of both the gold label and query, (2) \textit{SQL Inconsistency:} Generated samples are deemed inconsistent if one or more references fail to correctly reflect (semantically, logically, and technically) all key elements of the query (3) \textit{Gold-label Inconsistency:} Generated samples are marked inconsistent if one or more references do not correctly reflect (semantically and logically) all key elements of the gold-label, and (4) \textit{Full Consistency:} Generated samples are considered consistent when all references correctly reflect (semantically, logically, and technically) all key elements of both the gold label and query. 

\paragraph{Results and Validation}
We removed all inconsistent original reference utterances and SQL queries and generated captions to the SQL query, resulting in a final dataset of 853 SQL queries, 832 original utterances, and 2559 generated utterances (See \Cref{tab:final_dataset}). The evaluation revealed strong inter-rater reliability (Fleiss' Kappa $\kappa = 0.802$) and demonstrated that our iterative ICL prompt generates high-quality captions. This evaluation framework validates the quality of our generated dataset for advancing research in the field.

\section{RQ2: Graph-aware ICL Few-shot }

We now describe our proposed method and investigate a graph-aware ICL sample selection approach on the SQL2Text task. We developed two approaches for this task using the AST. Our first method, AST-ICL, leverages clustering with random selection from the cluster, and our second, AST-ICL-TOP, is a variant that selects the top $n$ similar examples. We compare our method to the following baselines (1) a random selection method, i.e., demonstration samples are selected using chance from the training set, and (2) a word overlap method, BM25.

\subsection{AST-ICL}
\begin{figure}
    \centering
    \includegraphics[width=.8 \linewidth]{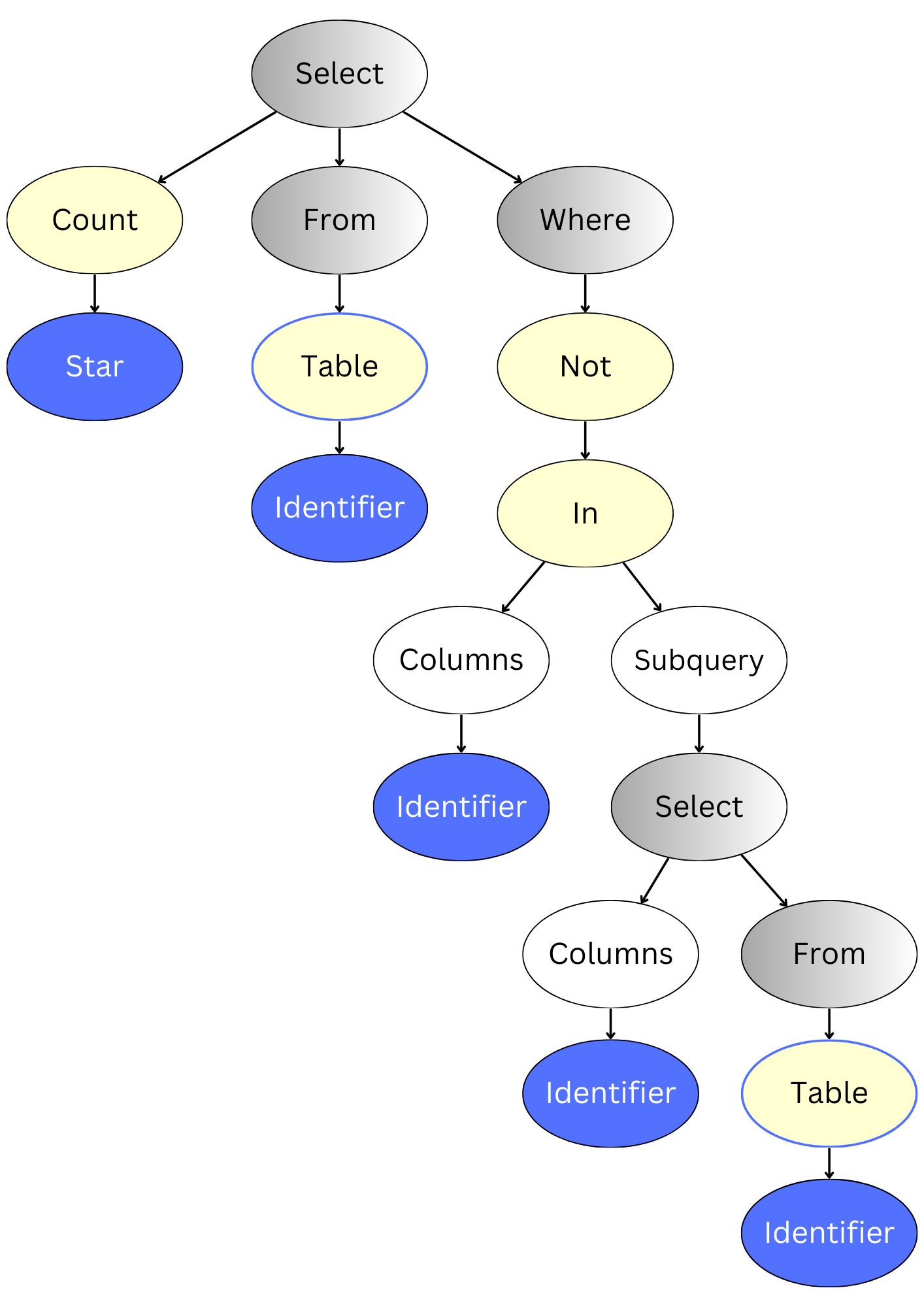}
    
    \caption{Parse Tree for SQL Query: \texttt{SELECT count(*) FROM Dogs WHERE dog\_id NOT IN (SELECT dog\_id FROM Treatments)}}
    \label{fig:graph_construction}
\end{figure}
\paragraph{AST} Our intuition for building a graph from the SQL for sample selection is based on the premise that the structure of the query (complex clauses, table joins, and nesting) is more informative than the word overlap. \Cref{fig:graph_construction} depicts the breakdown of an SQL query into a graph structure based on its AST. The AST provides a hierarchical structure that represents the syntactic structure of the SQL statement, such as keywords, operators, and operands.

\paragraph{Graph Construction}

Next, each node in the constructed AST graph is tokenized and embedded. This is followed by a Graph Convolution Network (GCN) aggregation steps:

\begin{equation}
    \mathbf{h}_i^{(l+1)} = \sigma \left( \sum_{j \in \mathcal{N}(i)} \frac{1}{\sqrt{d_i d_j}} \mathbf{W}^{(l)} \mathbf{h}_j^{(l)} \right)
\end{equation}
    where \( \mathcal{N}(i) \) is the set of neighbors of node \( i \), \( d_i \) and \( d_j \) are the degrees of nodes \( i \) and \( j \), \( \mathbf{W}^{(l)} \) is the weight matrix of layer \( l \), and \( \sigma \) is a non-linear activation function (e.g., ReLU).

To obtain a single vector representation for each AST, we perform graph pooling:
\begin{equation}
    \mathbf{h}_{\text{graph}} = \frac{1}{|V|} \sum_{i \in V} \mathbf{h}_i^{(L)}
\end{equation}
    where \( |V| \) is the number of nodes in the graph and \( \mathbf{h}_i^{(L)} \) is the node embedding at the final GCN layer \( L \).

The result of the graph pooling operation is a fixed-size embedding vector \( \mathbf{h}_{\text{graph}} \) for the entire AST, representing the SQL query in latent space.
\begin{figure*}
    \centering
    \includegraphics[width=1\linewidth]{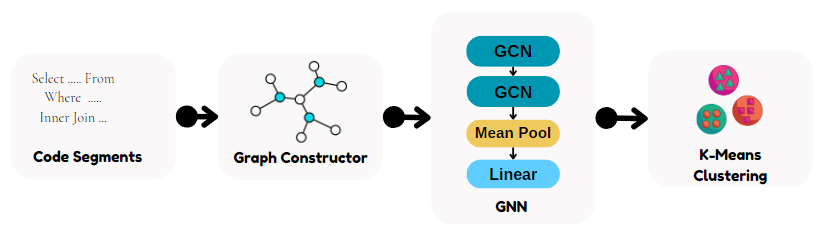}
    
    \caption{AST-ICL Model Overview: First, AST graphs are constructed for each SQL segment, then encoded into vectors by a GNN embedding model, and finally clustered using k-means for AST-based In-Context Learning (ICL).}
    \label{fig:graph}
\end{figure*}

Finally, we apply clustering to the graph embeddings to group similar SQL queries. During testing, the demonstration examples are picked randomly from the corresponding cluster. This improves the time complexity of the sample selection by statistical sampling from the corresponding cluster of the test query. We refer to the variation as AST-ICL (see~\Cref{fig:graph}).

\paragraph{AST-ICL-TOP} This variation follows the same approach of AST-ICL to generate the SQL AST-based graphs and embeddings. However, a full scan of the training set is used to select the  top $n$ similar samples. The computational requirement of this method is slightly higher since a full scan of the training set is performed to select the required samples instead of random selection from the corresponding cluster.

\subsection{Baselines}
We utilize two baselines: random selection, and BM25 word similarity. These two baselines establish a comprehensive evaluation framework to assess the effectiveness of our proposed AST-ICL sample selection method for ICL in SQL2Text tasks. The baselines are described in more detail in~\Cref{appen:baselines}.

\subsection{Evaluation Metrics}

To assess the performance of the AST-ICL in comparison to the baselines, we utilize three benchmark metrics: BERTScore, BLEU-4, and AlignScore.
BERTScore measures the similarity between the generated text in latent space. For BERTScore, we employ two fine-tuned transformer models optimized for paraphrasing: paraphrase-MiniLM-L6-v2 (BERTScore-1) and paraphrase-distilroberta-base-v1 (BERTScore-2)~\cite{reimers-2019-sentence-bert}. 

BLEU-4 calculates the geometric mean of n-gram precision scores (up to 4-grams) between generated text and reference, with a brevity penalty. It evaluates SQL-to-English translation quality by measuring n-gram overlap between generated text and reference questions.

AlignScore measures the logical consistency between the generated text and the reference utterances by evaluating the alignment of semantic content and reasoning patterns. Details of these metrics are provided in~\Cref{appen:eval metrics}.

\subsection{RQ2 Results}
First, we seek to understand whether our proposed method can outperform baseline few-shot technique. Second, we investigate LLM zero-shot capabilities. Third, we aim to understand how well open-source small-parameter LLMs compare to larger proprietary LLMs.

\subsubsection{AST-ICL vs. ICL few-shot}
We investigate the performance of our proposed method compared to the baseline few-shot methods. We examine the ability of AST-ICL to improve LLM semantic captioning capabilities and compare it to the current SOTA method.

\textbf{Our Graph-based method outperforms baseline ICL's few-shot sample selection for SQL semantic captioning}. \Cref{tab:rq11} presents the results for the benchmark datasets on the metrics. This experiment only used two demonstration samples for each method. This explains the higher-than-expected performance of random selection on some of the results. However, it is easily observable that AST-ICL-TOP achieves significantly higher performance than all other methods across all metrics.

\textbf{Additional samples significantly improve the SQL2Text task performance of AST-ICL methods}.  In the AST-ICL-TOP experiment, we measure the effect of adding additional samples on each method on the results. This is tabulated in \Cref{tab:rq12} for the CoSQL-S2T dataset for brevity. Additional results are reported in \Cref{appen:results}. Other methods also observe improvement, but to a lesser extent with random selection. These findings suggest that some samples may have limited positive influence or are redundant.
\begin{table}[h!]
   \centering
   \begin{adjustbox}{width=.5\textwidth,center}
       
   \begin{tabular}{l|l|ccr}
   \toprule
   \multirow{2}{*}{\textbf{Lang. Model}} & \multirow{2}{*}{\textbf{Method}} & \multicolumn{3}{c}{CoSQL-S2T Dataset} \\ 

   & & BERTScore-1 & BERTScore-2 & BLEU \\
   \midrule
      \multirow{1}{*}{GPT-4} 
        &  Zero-Shot 
& 0.8944 & 0.8724 & 0.4649\\

   \midrule
   \multirow{6}{*}{GPT-J}
        &  BM25-4
 & 0.8101 & 0.7767 & 0.2590 \\
        &  \textbf{AST-ICL-TOP-4} 
 & 0.8333 & 0.7974 & 0.2964 \\
        &  AST-ICL-4 
 & 0.8171 & 0.7778 & 0.2601 \\
        &  BM25-8 
 & 0.8240 & 0.7944 & 0.3000 \\
        &  \textbf{AST-ICL-TOP-8} 
 & 0.8350 & 0.7994 & 0.3125 \\
        &  AST-ICL-8 
 & 0.8294 & 0.7903 & 0.2785 \\

   \midrule
   \multirow{6}{*}{Mistral} 
        &  BM25-4
 & 0.8643 & 0.8298 & 0.3674 \\
        &  \textbf{AST-ICL-TOP-4} 
 & 0.8722 & 0.8331 & 0.3799 \\
        &  AST-ICL-4 
 & 0.8620 & 0.8340 & 0.3851 \\
        &  BM25-8 
 & 0.8785 & 0.8452 & 0.3960 \\
        &  \textbf{AST-ICL-TOP-8} 
 & 0.8722 & 0.8380 & 0.3988 \\
        &  AST-ICL-8 
 & 0.8720 & 0.8390 & 0.4112 \\

   \midrule
    \multirow{6}{*}{CodeLlama} 
        &  BM25-4
& 0.8478 & 0.8177 & 0.3566  \\
        &  \textbf{AST-ICL-TOP-4} 
& 0.8667 & 0.8320 & 0.3560  \\
        &  AST-ICL-4 
& 0.8557 & 0.8311 & 0.3390  \\
        &  BM25-8 
& 0.8664 & 0.8382 & 0.3935  \\
        &  \textbf{AST-ICL-TOP-8} 
& 0.8699 & 0.8327 & 0.3709  \\
        &  AST-ICL-8 
& 0.8658 & 0.8343 & 0.3611  \\
   \bottomrule

   \end{tabular}
   \end{adjustbox}
    \caption{Test Results for competitive small-parameter LLMs vs GPT-4 Zero-shot}
   \label{tab:rq3}
\end{table}
\subsubsection{Zero-shot Vanilla v.s. Few-shot }
We investigate the performance of zero-shot methods compared to baseline few-shot method. We compare the capacity of LLMs to understand and translate SQL with no demonstration sample.

\textbf{The performance of the models on zero-shot was quite poor.} A cursory review of the output suggests that some results included or consisted entirely of more SQL. Different prompt restrictions were used to avoid additional code generation. A key finding is that while LLMs tested performed poorly in zero-shot, experiments suggest that only a small number of ICL samples can improve the result (\Cref{tab:rq11}).


\subsubsection{Smaller vs. Larger LLM}
We investigate the performance of three pre-trained smaller-sized LLMs: GPT-J, Mistral, and CodeLlama. We utilize GPT-4 as our larger-sized LLM, given its superior performance across NLP tasks. GPT-4  is reported to have 100 trillion parameters. This is about 14k times larger than the small models considered in this work. Here we attempt to measure the capacity of GPT-4 to understand SQL code, and compare the performance gains achieved from AST-ICL with GPT-4. \Cref{prompt:GPT-4} includes the prompt used with GPT-4, which was verified using OpenAI playground to generate the intended results.

GPT-4 was shown to perform relatively well on the SQL2Text task using zero-shot. Here we investigate if the performance of GPT-4 can be improved with ICL, and whether AST-ICL would outperform a random selection in this setting (\Cref{tab:rq4}).


\textbf{GPT-4 shows some performance improvement using ICL}. This finding demonstrates that ICL do not contribute to helping large LLMs as well as smaller LLMs in improving code understanding (\Cref{tab:rq4}). However, there is a notable improvement in BLEU score.

\textbf{There is no clear advantage in using AST-ICL sample selection vs. other approaches using GPT-4}. Larger models understand SQL very well in zero-shot, so adding ICL does not add any significant improvement (\Cref{tab:rq4}).
 
\begin{table}[h!]
   \centering
   \begin{adjustbox}{width=.5\textwidth,center}

   \begin{tabular}{l|ccc}
   \toprule
    \multirow{2}{*}{\textbf{Method}} & \multicolumn{3}{c}{CoSQL-S2T Dataset} \\ 

   &  BERTScore-1 & BERTScore-2 & BLEU \\
   \midrule
     GPT-4 Zero-shot & 0.8944 & 0.8724 & 0.4649 \\

   \midrule

  GPT-4 + BM25-2 & 0.9035 & 0.8655 & 0.5405 \\
  GPT-4 + \textbf{AST-ICL-TOP-2} & 0.8999 & 0.8652 & 0.5107 \\
  GPT-4 + Random-2 & 0.9017 & 0.8701 & 0.5249 \\

   \bottomrule

   \end{tabular}
   \end{adjustbox}
    \caption{Test Results for GPT-4 using different ICL prompts}
   \label{tab:rq4}
\end{table}




\section{Discussion}
\textbf{AST-ICL demonstrates a remarkable capacity to improve the effectiveness of smaller LLMs for semantic captioning}. Our graph-aware methods for SQL2Text outperform alternative sample selection methods for ICL, such as BM25, while providing more efficiency. In particular, the capacity to choose more samples achieved performance slightly lower than GPT-4 from smaller models by orders of magnitude. The GNN-based encoding method of AST-ICL and AST-ICL-TOP using the syntactical elements of SQL based on the AST indicates that LLMs exhibit enhanced reasoning ability when the examples are structurally similar, compared to word similarity in most cases.

\textbf{Smaller LLMs need ICL for semantic captioning}. Even though SQL is a popular language incorporated in the training parameter of smaller LLMs, ICL is integral to achieving decent performance on SQL2Text. This suggests that smaller LLMs cannot follow prompt instructions very well for this task, unlike GPT-4, which can perform the task correctly in zero-shot. 

\textbf{Random ICL quickly suffers from diminishing returns}. An ICL sample selection method based on a strategy, even a simple one, appears to perform better with more samples. However, random selection can perform well with a few samples (e.g. 2 samples), and outperform BM25 on some metrics. This suggests that word similarity may be counter-productive for small sample sizes, and may provide limited diversity.

\textbf{AST-ICL improves GPT-4, but insignificantly}. While AST-ICL-TOP improves GPT4, our comparative analysis reveals important insights about the relationship between model size and semantic captioning performance. \Cref{tab:rq3}  shows GPT-4, with approximately 100 trillion parameters, achieves a zero-shot BLEU score of 0.4649 on the CoSQL-S2T dataset. However, smaller models enhanced with AST-ICL demonstrate competitive performance. Specifically, Mistral-7B with AST-ICL achieves a BLEU score of 0.4112, maintaining a comparative performance to GPT-4 while requiring significantly fewer computational resources. The efficiency advantages of smaller models become particularly evident in production environments. This efficiency-performance trade-off provides a compelling argument for using smaller models in practical applications.

\section{Conclusion} Our work introduces the problem of semantic captioning as the inverse problem of the well-known semantic parsing downstream task, focusing on smaller parameter LLMs, which may be easier to integrate into coding and educational platforms. In order to motivate additional work on this problem, we have provided a method to repurpose, generate, and publish new datasets, and provide an iterative prompt for the community to reuse for additional datasets.

Furthermore, we demonstrate that for the SQL2Text task of semantic parsing, leveraging the graph properties of SQL can provide added value. These findings have the potential to facilitate the integration of smaller LLMs into coding and educational platforms, making them more accessible and efficient for a wide range of users. Moreover, our work contributes to the broader understanding of LLMs’ capabilities in code comprehension and generation tasks, paving the way for further research and applications in this domain.






\section{Limitations and Future Work}

\subsection{Current Limitations}
Our work has several key limitations that warrant consideration. First, while our approach demonstrates effectiveness with SQL, its applicability may vary across different programming languages and domains. Second, although we utilize established benchmarks, our current datasets may not fully capture the complexity and diversity of real-world SQL queries and their natural language descriptions. In our preliminary experiments with datasets like KaggleDBQA, we encountered excessive inconsistencies that necessitated excluding them from our analysis, highlighting the challenges in dataset quality and curation.

\subsection{Future Research Directions}

Several important research directions emerge from our work. First, our experience with inconsistent datasets underscores the critical need for robust quality metrics in SQL-text paired datasets. Future work should focus on systematic evaluation of existing benchmarks and creation of new, high-quality datasets that better reflect real-world complexity. The development of standardized quality assessment protocols will be essential for advancing the field.

In terms of technical advancements, as the field evolves, research should investigate the scaling behavior of AST-ICL with larger models and explore potential hybrid approaches combining different model sizes. The development of more sophisticated graph representation techniques will be crucial, as will extending our approach to other programming languages and domains. These technical improvements could significantly enhance the versatility and effectiveness of semantic captioning systems.

Additionally, the evaluation framework for semantic captioning systems requires substantial development. Future research should expand evaluation criteria to incorporate domain-specific requirements while developing standardized quality assessment protocols. Incorporating more comprehensive human evaluation methodologies will be crucial for validating system performance across diverse use cases. These enhancements to evaluation frameworks will provide more reliable insights into system performance and further advance semantic captioining.

These directions could potentially improve performance while maintaining computational efficiency. The emergence of more powerful LLMs presents opportunities to push performance boundaries further, particularly in addressing the dataset quality challenges we encountered. Additionally, expanding the evaluation framework to include more diverse SQL queries and domain-specific applications would provide valuable insights for practical deployments.

\section*{Acknowledgments}

The authors sincerely thank Rupak Kumar Das, a PhD student, for participating in the human evaluation, which significantly contributed to the quality of this study. We also extend our gratitude to Dr. Adaku Uchendu, AI Researcher at MIT Lincoln Lab, for her important guidance and support throughout this work.
\bibliography{custom}

\appendix

\section{Supplement on Experimental Setting}
\label{sec:appendix}

\subsection{Description of Datasets}
 The SQL2Text datasets we used are as follows:

\textbf{CoSQL} \cite{yu2019cosql}: A large-scale dataset for cross-domain, general-purpose database querying dialogue systems, consisting of 30k+ turns with 10k+ annotated SQL queries from 3k dialogues querying 200 databases across 138 domains. 

\textbf{Spider} \cite{yu2018spider}: A large-scale, complex, and cross-domain semantic parsing and Text2SQL dataset with 10K+ questions and 5K+ unique complex SQL queries on 200 databases covering 138 domains, focusing on complex SQL queries (JOIN, GROUP BY, nested queries) and requiring generalization to new queries and schemas.  

\textbf{SParC} \cite{yu2020sparc}: A dataset for cross-domain Semantic Parsing in Context, consisting of 4K+ coherent question sequences (12k+ questions with SQL queries) from controlled user interactions with 200 databases over 138 domains, demonstrating complex contextual dependencies and requiring generalization to new domains. 

\begin{table}
  \centering
  \scalebox{0.65}{
    \begin{tabular}{c|c|c|ccc|ccc}
\toprule
        \multirow{2}{*}{\textbf{Dataset}} & 
        \multirow{2}{*}{\textbf{Count}} &  
        \multirow{2}{*}{\textbf{Type}} & \multicolumn{3}{c}{Query Length} & 
        \multicolumn{3}{c}{Keywords} \\
        &&
        & min & max & mean 
        & min & max & mean \\
        \midrule
         \multirow{2}{*}{CoSQL} & 2159 & Train & 6 & 87 & 19.7 & 2 & 14 & 5.5\\
         & 293 & Val & 6 & 58 & 19.8 & 2 & 12 & 5.5\\
        \midrule
        \multirow{2}{*}{SParC} & 3034 & Train & 4 & 87 & 17.6 & 2 & 14 & 5.0 \\
        & 422 & Val & 4 & 58 & 17.5 & 2 & 13 & 5.0\\
        \midrule
        \multirow{2}{*}{Spider} & 7000 & Train & 4 & 87 & 15.9 & 2 & 14 & 4.6 \\
         & 1034 & Val & 4 & 58 & 15.6 & 2 & 13 & 4.6\\
        \bottomrule
    \end{tabular}
    }
    \caption{Statistics of query length and SQL keyword occurrence (e.g. \texttt{SELECT, FROM}) in training and validation samples across different datasets}
    \label{table:samples}
\end{table}

\subsection{Evaluation Metrics}
\label{appen:eval metrics}
We describe the evaluation metrics we employed for assessing the quality of generated utterances and for our semantic caption generation task
\begin{itemize}
    \item BERTScore \cite{reimers-2019-sentence-bert}: we employ two transformer variants fine-tuned for paraphrasing on different general datasets: (1) paraphrase-MiniLM-L6-v2\footnote{\url{https://huggingface.co/sentence-transformers/paraphrase-MiniLM-L6-v2}} (BERTScore-1) and (2) paraphrase-distilroberta-base-v1\footnote{\url{https://huggingface.co/sentence-transformers/paraphrase-distilroberta-base-v1}} (BERTScore-2). We selected these fine-tuned transformer variants given their optimization for paraphrasing and their strong performance on semantic search tasks. BERTScore measures the similarity between the generated text and the original prompt or question using their corresponding transformer embeddings.
    \item BLEU-4 \cite{papineni2002bleu}: we employ BLEU-4 (BLEU), which calculates the geometric mean of n-gram precision scores (up to 4-grams) between the generated text and the reference, multiplied by a brevity penalty to penalize short translations. BLEU evaluates the quality of the translation from SQL to English by measuring the overlap of n-grams between the generated text and the reference question.
    
    \item AlignScore \cite{zha2023alignscore}: we employ AlignScore, a factual alignment or logical consistency evaluation metric designed to measure how well the generated text aligns with the original source material in terms of meaning and factual correctness. It compares factual statements in the generated text with the source to assess consistency. This metric is particularly useful for evaluating tasks like text summarization and semantic captioning while preserving original meaning and facts.

\end{itemize}

\subsection{LLMs} 
\label{appen:LLMs}
We leverage several LLMs of varying sizes for two primary purposes: (1) the generation of utterances to develop a semantic caption dataset using SQL queries from benchmark Text2SQL corpora and (2) the generation of semantic captions using our proposed AST-ICL and baseline methods.

\noindent We employ the following LLM for semantic caption generation:
\begin{itemize}
    \item GPT-4o: A multimodal model that accepts text or image inputs and generates text outputs. It offers comparable intelligence to GPT-4 Turbo, but with twice the generation speed and 50\% lower cost. GPT-4o also provides superior vision capabilities and excels in non-English languages, making it highly efficient for tasks like semantic captioning.
\end{itemize}

\noindent We utilize the following LLMs for the evaluation of generated semantic captions:
\begin{itemize}
    \item GPT-J-6B \cite{gpt-j}: A GPT-based LLM developed by EleutherAI, enhanced for complex NLP generation and understanding applications\footnote{\href{https://huggingface.co/EleutherAI/gpt-j-6b}{GPT-J Huggingface: EleutherAI/gpt-j-6b}}.
    \item Mistral-7B \cite{jiang2023mistral}: A transformer-based LLM optimized for large-scale language understanding and generation tasks\footnote{\href{https://huggingface.co/mistralai/Mistral-7B-v0.1}{Mistral Huggingface: mistralai/Mistral-7B-v0.1}}.
    \item CodeLlama-7B \cite{roziere2023code}: A custom transformer architecture developed by Meta AI, efficient and scalable for a variety of NLP tasks, with a recent focus on code understanding and generation\footnote{\href{https://huggingface.co/meta-llama/CodeLlama-7b-hf}{CodeLlama Huggingface: meta-llama/CodeLlama-7b-hf}}.
    \item GPT-4 \cite{achiam2023gpt}: OpenAI's advanced multimodal LLM, excelling in text generation, image processing, and diverse tasks including natural language understanding, programming, and problem-solving.
\end{itemize}

\subsection{ICL Prompt Selection Baselines}\label{appen:baselines}
Here, we provide additional details on the two baselines used in our experiments. 
\paragraph{Random Sampling}
This approach is a simple baseline where demonstration samples are randomly selected from the training set without considering any specific criteria or relevance to the target example. Despite its simplicity and low computational cost, random sampling may not always select the most informative or relevant samples. By comparing our proposed method to random sampling, we can assess whether our approach significantly improves over a simple and unbiased selection strategy.

\paragraph{BM25}
 This is a prominent word overlap-based method widely used in information retrieval tasks. It calculates the relevance score between a query and a document based on the frequency and importance of the query terms appearing. In the context of sample selection for ICL, BM25 selects demonstration samples with a high word overlap with the target example, aiming to select the most relevant samples based on their textual similarity. By comparing our proposed method to BM25, we can evaluate whether our AST-ICL approach provides additional benefits over a word overlap-based method in the context of SQL2Text tasks.

\subsection{Details of Experimental Cost}
The costs of running the GPT-4 Experiments are quantified in \Cref{tab:cost}. In addition, the total cost of running GPT-4o generation of Iterative ICL datasets was \$6.87 almost evenly split across the three datasets.

\begin{table}[h!]
   \centering
   \scalebox{.9}{
   \begin{tabular}{l|r}
   \toprule
     \multicolumn{2}{c}{GPT-4 Experiments} \\ 

   \midrule
     GPT-4 Zero-shot & 1.51  \\

   \midrule

  GPT-4 + BM25-2 & 1.79  \\
  GPT-4 + \textbf{AST-ICL-TOP-2} & 1.68  \\
  GPT-4 + Random-2 & 1.89 \\
 
   \midrule
GPT-4o Datasets Generation & 6.86\\
\midrule
\textbf{Total} & \$13.73 \\
\bottomrule

   \end{tabular}
   }
    \caption{Total USD Cost of GPT-4 Experiments}
    \label{tab:cost}
\end{table}

The prompt used for the ICL experiments run on GPT-4 is as follows:
\begin{tcolorbox}[colback=blue!10, colframe=blue!40!black, title=GPT-4 Prompt, rounded corners]
\textbf{System}: "You are an intelligent SQL Code assistant who effectively translates the intent and logic of the SQL queries into natural language that is easy to understand.
\textbf{User}: Convert the given SQL query into a clear and concise natural language query limited to 1 sentence.  Ensure that the request accurately represents the actions specified in the SQL query and is easy to understand for someone without technical knowledge of SQL.
\label{prompt:GPT-4}
\end{tcolorbox}

\subsection{Experimental Environment}
The sqlglot\footnote{\url{https://github.com/tobymao/sqlglot}} Python library was used to parse SQL statements. We used Python 3.11.8 and PyTorch 2.2.1 with CUDA v12.4, and PyTorch Geometric 2.5.3. All experiments were run on (1) NVIDIA A100 32GB GPU. 

\subsection{Implementation Details}
\paragraph{Tokenizer} A simple index tokenizer converts the SQL syntax tree node elements to a numeric form. Torch \texttt{nn.Embedding} was used to embed the token to a vector of 100. The edge index was constructed as directed edges as depicted in~\Cref{fig:graph_construction}.

Two GCN aggregation layers were used to aggregate the node embeddings followed by a global mean pool and a linear layer into vector outputs in a 2-dim space. For the AST-ICL variation, the vector outputs are clustered using k-means with 20 clusters. The number of clusters was configured based on the Silhouette score.

\begin{figure}
    \centering
    \includegraphics[width=.9 \linewidth]{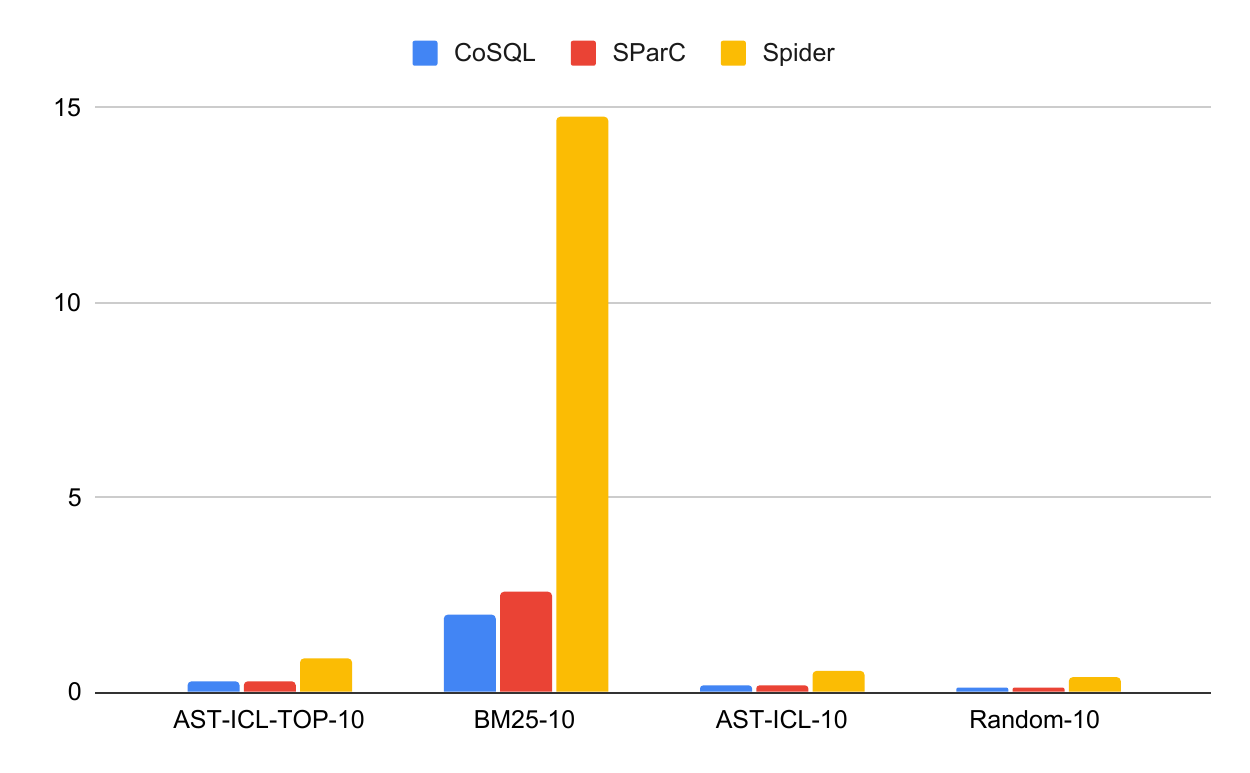}
    
    \caption{The total time (s) to generate the prompt samples for ICL by method, and dataset. Smaller datasets demonstrate a shorter time}
    \label{fig:generationtime}
\end{figure}
\subsection{RQ1 Results}\label{appen:results1}
\begin{table}
   \centering
 
   \begin{tabular}{l|c}
   \toprule
     Dataset & Count \\ 

   \midrule

  CoSQL-S2T & 52  \\
  \midrule
  SparC-S2T & 68  \\
  \midrule
  Spider-S2T & 65 \\
 
   \bottomrule
   \end{tabular}
    \caption{The number of utterance improvements between the initial and the final generation using Iterative ICL.}
    \label{tab:alignscore}
\end{table}
\Cref{tab:alignscore} outlines the number of utterances improved using Iterative ICL's approach based on the generated and final refined variations.

\Cref{tab:old-vs-new} shows the effect on the measured scores when the newly generated datasets are used vs. using the datasets in their original form.

\begin{table}[h!]
   \centering
   \begin{adjustbox}{width=.5\textwidth,center}
   \begin{tabular}{ll|cccc}
   \toprule
    \multirow{2}{*}{LLM} & \multirow{2}{*}{Dataset} & \multicolumn{3}{c}{\textbf{AST-ICL-TOP-2} Results} \\ 

   & & BERTScore-1 & BERTScore-2 & BLEU \\
   \midrule
     \multirow{2}{*}{GPT-J} 
& CoSQL-S2T  & 0.8153 & 0.7783 & 0.2813\\
& CoSQL  & 0.7584 & 0.7135 & 0.1914\\
   \midrule
     \multirow{2}{*}{Mistral}  
& CoSQL-S2T & 0.8632 & 0.8306 & 0.3732\\
& CoSQL & 0.8194 & 0.7668 & 0.2417\\
   \midrule
    \multirow{2}{*}{CodeLlama} 
& CoSQL-S2T & 0.8571 & 0.8217 & 0.3578\\
& CoSQL & 0.7933 & 0.7407 & 0.2012\\
\bottomrule
   \end{tabular}
\end{adjustbox}
    \caption{Test Results for AST-ICL-TOP-2 using original CoSQL and newly-introduced CoSQL-S2T dataset for SQL2Text}
   \label{tab:old-vs-new}
\end{table}

\subsection{RQ2 Results}\label{appen:results}

\Cref{tab:rq12} demonstrates the effect of the number of demonstration samples on the results. \Cref{fig:generationtime} outlines the training time requirements for the various methods used.

\begin{table*}[t!]
   \centering
 \begin{adjustbox}{width=1\textwidth,center}

   \begin{tabular}{l|l|ccrr|ccrr|ccrr}
   \toprule
   \multirow{2}{*}{Samples Size}& 
   \multirow{2}{*}{Method}& 
   \multicolumn{4}{c|}{GPT-J} &  
   \multicolumn{4}{c|}{Mistral}& 
   \multicolumn{4}{c}{CodeLlama}   \\ 
   & &
    BERTScore-1 & BERTScore-2 & BLEU & AlignScore &
    BERTScore-1 & BERTScore-2 & BLEU & AlignScore &
    BERTScore-1 & BERTScore-2 & BLEU & AlignScore\\
   \midrule
   \multirow{4}{*}{Two}  
&  Random-2
& 0.7837 & 0.7392 & 0.2021 & 0.4943
& 0.8515 & 0.8145 & 0.3509 & 0.7335
& 0.8384 & 0.8019 & 0.2996 & 0.6756
\\
&  BM25-2  
& 0.7779 & 0.7421 & 0.2197 & 0.5189
& 0.8498 & 0.8143 & 0.3415 & 0.7499
& 0.8350 & 0.8022 & 0.3314 & 0.7153
 \\
&  \textbf{AST-ICL-TOP-2} 
& \textbf{0.8153}$^{\dagger\ddagger}$ & \textbf{0.7783}$^{\dagger\ddagger}$ & \textbf{0.2813}$^{\dagger\ddagger}$ & \textbf{0.6357}$^{\dagger\ddagger}$
& \textbf{0.8632} & \textbf{0.8306}$^{\dagger\ddagger}$ & \textbf{0.3732} & \textbf{0.8103}$^{\dagger\ddagger}$
& \textbf{0.8571}$^{\dagger\ddagger}$ & \textbf{0.8217}$^{\dagger\ddagger}$ & \textbf{0.3578}$^\dagger$ & \textbf{0.7766}$^{\dagger\ddagger}$
 \\
&  \textbf{AST-ICL-2} 
& 0.8082$^{\dagger\ddagger}$ & 0.7711$^{\dagger\ddagger}$ & 0.2376 & 0.5583
& 0.8577 & 0.8200 & 0.3369 & 0.7382
& 0.8451 & 0.8085 & 0.3218 & 0.7031
\\
   \midrule
   \multirow{4}{*}{Four} 
&  Random-4
& 0.8032 & 0.7660 & 0.2424 & 0.5247
& 0.8663 & 0.8288 & 0.3646 & 0.7474
& 0.8534 & 0.8209 & 0.3310 & 0.7185
\\
&  BM25-4  
& 0.8101 & 0.7767 & 0.2590 & 0.5671
& 0.8643 & 0.8298 & 0.3674 & 0.7767
& 0.8478 & 0.8177 & \textbf{0.3566} & 0.7402
 \\
&  \textbf{AST-ICL-TOP-4} 
& \textbf{0.8333}$^{\dagger\ddagger}$ & \textbf{0.7974}$^{\dagger\ddagger}$ & \textbf{0.2964}$^\dagger$ & \textbf{0.6457}$^{\dagger\ddagger}$
& \textbf{0.8722} & \textbf{0.8331} & \textbf{0.3799} & \textbf{0.7978}
& \textbf{0.8667}$^\ddagger$ & \textbf{0.8320} & 0.3560 & \textbf{0.7865}$^\dagger$
 \\
&  \textbf{AST-ICL-4} 
& 0.8171 & 0.7778 & 0.2601 & 0.5745
& 0.8620 & 0.8340 & 0.3851 & 0.7674
& 0.8557 & 0.8311 & 0.3390 & 0.7395
\\
   \midrule
   \multirow{4}{*}{Eight} 
&  Random-8
& 0.8176 & 0.7813 & 0.2462 & 0.5955
& 0.8717 & 0.8388 & 0.3952 & 0.7949
& 0.8600 & 0.8278 & 0.3623 & 0.7481
\\
&  BM25-8  
& 0.8240 & 0.7944 & 0.3000$^\dagger$ & 0.5748
& 0.8785 & 0.8452 & 0.3960 & 0.8362
& 0.8664 & \textbf{0.8382}& \textbf{0.3935} & \textbf{0.7888}
 \\
&  \textbf{AST-ICL-TOP-8} 
& \textbf{0.8350}$^\dagger$ & \textbf{0.7994}$^\dagger$ & \textbf{0.3125}$^\dagger$ & \textbf{0.6272}
& 0.8722 & 0.8380 & \textbf{0.3988} & 0.7958
& \textbf{0.8699} & 0.8327 & 0.3709 & 0.7807
 \\
&  \textbf{AST-ICL-8} 
& 0.8294 & 0.7903 & 0.2785 & 0.6079
& 0.8720 & 0.8390 & 0.4112 & 0.7868
& 0.8658 & 0.8343 & 0.3611 & 0.7745
\\
\bottomrule
   \end{tabular}
   \end{adjustbox}
   \caption{Results against different demonstration sample sizes using the CoSQL-S2T dataset. The number after the hyphen refers to the number of demonstration samples used in the prompt, which is two in this experiment. $^\dagger$ denotes t-test statistical significance compared to Random. $^\ddagger$ denotes t-test statistical significance compared to BM25.}
   \label{tab:rq12}
\end{table*}

\subsection{Iterative ICL Prompt}\label{fig:CoC_prompt}
The Iterative ICL prompt used to generate the new datasets is outlined in \Cref{fig:gen-prompt} below.

\subsection{Additional Experiments}
\Cref{tab:gnn-type} presents results based on  GNN architecture. Our choice of GCN is motivated by its AlignScore results on which it has a slight edge (5 out of 9) compared to other options. As future work, it would be interesting to explore the other architectures comprehensively.

\Cref{tab:query-type} presents the results based on the SQL query type. While nested and aggregate queries have lower zero-shot results, they benefit significantly from a higher sample size. In addition, AST-ICL provides the most improvement gains on these query types.

\begin{figure*}[p]
\begin{tcolorbox}[colback=blue!10, colframe=blue!40!black, title=Chain of Commentation Prompt, rounded corners]
\scriptsize
\textbf{System Prompt}: 
Given a user SQL query, follow these 3 steps to generate, review, and refine variations of utterances or questions that can be used to create the query accurately in a text-to-SQL generation task.\\

    Step 1: Generate three utterances or question variations for the SQL query
    
    - Produce three different but accurate variations of utterances or questions for the user’s original SQL query.
    
    - Ensure each variation presents a unique phrasing while maintaining the original query's intent, logic, key elements, and meaning.

    Example:
\begin{verbatim}
{
    "Original SQL Query": "SELECT * FROM Flights WHERE city_from = 'New York' AND city_to = 'Boston' ORDER BY price ASC LIMIT 1;",
    "Generated Variations": [
        "What is the most affordable flight from New York to Boston?",
        "Can you show me the cheapest flight available between New York and Boston?",
        "Find the lowest-priced flight from New York to Boston."
    ]
}
\end{verbatim}
In Step 2, review the generated variations by identifying any errors, ambiguities, or areas for improvement in each variation. Focus on accuracy, clarity, and alignment with the original SQL query's intent, and provide feedback for each variation.

Example:
\begin{verbatim}
{
    "Review Feedback": {
        "Variation 1": {
            "Feedback": "This variation is clear but could explicitly mention it is a search for a flight."
        },
        "Variation 2": {
            "Feedback": "Good phrasing, but it could specify 'currently available' for more precision."
        },
        "Variation 3": {
            "Feedback": "Accurate and concise. No changes needed."
        }
    }
}
\end{verbatim}

In Step 3, refine and finalize the utterances by improving each variation based on the feedback in Step 2, ensuring better accuracy, clarity, fluency, and relevance to the original SQL query. Finalize the three refined versions of the utterances.

Example:
\begin{verbatim}
{
    "Final Refined Variations": [
        "What is the most affordable flight currently available from New York to Boston?",
        "Can you show me the cheapest flight available between New York and Boston?",
        "Find the lowest-priced flight from New York to Boston in the current listings."
    ]
}
\end{verbatim}

Now, fill out the form below based on the Original SQL Query input from the user:
\begin{verbatim}
{
    "Original SQL Query": "<User's SQL Query Input>",
    "Generated Variations": [
        "<Generated Variation 1>",
        "<Generated Variation 2>",
        "<Generated Variation 3>"
    ]
}
\end{verbatim}

For Step 2:
\begin{verbatim}
{
    "Review Feedback": {
        "Variation 1": {
            "Feedback": "<Feedback for Variation 1>"
        },
        "Variation 2": {
            "Feedback": "<Feedback for Variation 2>"
        },
        "Variation 3": {
            "Feedback": "<Feedback for Variation 3>"
        }
    }
}
\end{verbatim}

For Step 3:
\begin{verbatim}
{
    "Final Refined Variations": [
        "<Refined Variation 1>",
        "<Refined Variation 2>",
        "<Refined Variation 3>"
    ]
}
\end{verbatim}
\end{tcolorbox}
\label{fig:gen-prompt}
\end{figure*}

\begin{table*}[t!]
   \centering
 \begin{adjustbox}{width=1\textwidth,center}

   \begin{tabular}{l|l|ccrr|ccrr|ccrr}
   \toprule
   \multirow{2}{*}{Samples Size}& 
   \multirow{2}{*}{Method}& 
   \multicolumn{4}{c|}{Simple (129 instances)} &  
   \multicolumn{4}{c|}{Nested (42 instances)} & 
   \multicolumn{4}{c}{Aggregate (119 instances)}   \\ 
   & &
    BERTScore-1 & BERTScore-2 & BLEU & AlignScore &
    BERTScore-1 & BERTScore-2 & BLEU & AlignScore &
    BERTScore-1 & BERTScore-2 & BLEU & AlignScore\\
   \midrule
   Zero 
& Zero-Shot
& 0.5994 & 0.5125 & 0.0115 & 0.5140
& 0.5718 & 0.4795 & 0.0065 & 0.2258
& 0.5693 & 0.4679 & 0 & 0.4804 \\
      \midrule

   \multirow{4}{*}{Two}  
&  Random-2 

& 0.7660 & 0.7216 & 0.1740 & 0.4306
& 0.7578 & 0.7190 & 0.1645 & 0.4452
& 0.8119 & 0.7654 & 0.2458 & 0.5807

\\
&  BM25-2  
& 0.7453 & 0.7141 & 0.1612 & 0.4285
& 0.8118$^\dagger$ & 0.7791$^\dagger$ & 0.2735$^\dagger$ & 0.5747
& 0.8013 & 0.7594 & 0.2642 & 0.5971
 \\
&  \textbf{AST-ICL-TOP-2} 
& \textbf{0.8040}$^{\dagger\ddagger}$ & \textbf{0.7799}$^{\dagger\ddagger}$ & \textbf{0.2463}$^{\dagger\ddagger}$ & \textbf{0.6052}$^{\dagger\ddagger}$
& \textbf{0.8438}$^\dagger$ & \textbf{0.8107}$^\dagger$ & \textbf{0.3572}$^\dagger$ & \textbf{0.7186}$^\dagger$
& \textbf{0.8300} & \textbf{0.7981}$^{\dagger\ddagger}$ & \textbf{0.2959} & \textbf{0.7020}$^{\dagger\ddagger}$

\\
&  \textbf{AST-ICL-2} 
& 0.7963$^{\dagger\ddagger}$ & 0.7536$^{\dagger\ddagger}$ & 0.2088 & 0.5428$^{\dagger\ddagger}$
& 0.8098$^\dagger$ & 0.7828$^\dagger$ & 0.2488 & 0.4666
& 0.8204 & 0.7859 & 0.2649 & 0.6076
 \\
   \midrule
   \multirow{4}{*}{Four} 
&  Random-4
& 0.7733 & 0.7465 & 0.1985 & 0.4570
& 0.8029 & 0.7596 & 0.2381 & 0.5299
& 0.8357 & 0.7893 & 0.2916 & 0.5963
\\
&  BM25-4  
& 0.7916 & 0.7665 & 0.2189 & 0.5252
& 0.8274 & 0.7996 $^\dagger$ & 0.2720 & 0.5711
& 0.8241 & 0.7796 & 0.2980 & 0.6111
 \\
&  \textbf{AST-ICL-TOP-4} 
& \textbf{0.8145}$^\dagger$ & \textbf{0.7830}$^\dagger$ & \textbf{0.2483} & \textbf{0.5750}$^\dagger$
& \textbf{0.8435}$^\dagger$ & \textbf{0.8039}$^\dagger$ & \textbf{0.3038} & \textbf{0.6529}
& \textbf{0.8501}$^\ddagger$ & \textbf{0.8107}$^\ddagger$ & \textbf{0.3461} & \textbf{0.7197}$^{\dagger\ddagger}$
 \\
&  \textbf{AST-ICL-4} 
& 0.7959 & 0.7606 & 0.2085 & 0.4918
& 0.8280 & 0.8044$^\dagger$ & 0.2984 & 0.6155
& 0.8362 & 0.7871 & 0.3026 & 0.6496
\\
   \midrule
   \multirow{4}{*}{Eight} 
&  Random-8
& 0.7963 & 0.7669 & 0.2234 & 0.5451
& 0.8213 & 0.7892 & 0.2782 & 0.5667
& 0.8395 & 0.7941 & 0.2597 & 0.6603
\\
&  BM25-8  
& 0.8054 & 0.7788 & 0.2521 & 0.5057
& \textbf{0.8513} & \textbf{0.8273} & \textbf{0.3627} & 0.6204
& 0.8345 & \textbf{0.7998} & \textbf{0.3298} & 0.6337
 \\
&  \textbf{AST-ICL-TOP-8} 
& \textbf{0.8246}$^\dagger$ & \textbf{0.7971}$^\dagger$ & \textbf{0.3007}$^\dagger$ & \textbf{0.5693}
& 0.8463 & 0.8175 & 0.3270 & \textbf{0.6227}
& \textbf{0.8423} & 0.7956 & 0.3201 & \textbf{0.6915}
 \\
&  \textbf{AST-ICL-8} 
& 0.8104 & 0.7774 & 0.2501 & 0.5403
& 0.8380 & 0.8048 & 0.3133 & 0.5784
& 0.8470 & 0.7991 & 0.2971 & 0.6915
\\
\bottomrule
   \end{tabular}
   \end{adjustbox}
   \caption{Results based on query type and number of demonstrations on GPT-J using the CoSQL-S2T dataset. The number after the hyphen refers to the number of demonstration samples used in the prompt. $^\dagger$ denotes t-test statistical significance compared to Random. $^\ddagger$ denotes t-test statistical significance compared to BM25.}
   \label{tab:query-type}
\end{table*}

\begin{table*}[t!]
   \centering
 \begin{adjustbox}{width=1\textwidth,center}

   \begin{tabular}{l|l|ccrr|ccrr|ccrr}
   \toprule
   \multirow{2}{*}{Samples Size}& 
   \multirow{2}{*}{Graph Method}& 
   \multicolumn{4}{c|}{GPT-J} &  
   \multicolumn{4}{c|}{Mistral} & 
   \multicolumn{4}{c}{CodeLlama}   \\ 
   & &
    BERTScore-1 & BERTScore-2 & BLEU & AlignScore &
    BERTScore-1 & BERTScore-2 & BLEU & AlignScore &
    BERTScore-1 & BERTScore-2 & BLEU & AlignScore\\
   \midrule

   \multirow{3}{*}{Two}  
&  GCN 
& 0.8153 & 0.7783 & 0.2813 & 0.6357
& 0.8632 & 0.8306 & 0.3732 & 0.8103
& 0.8571 & 0.8217 & 0.3578 & 0.7766

\\
&  GraphSAGE  
& 0.8239 & 0.7896 & 0.2820 & 0.6307
& 0.8711 & 0.8409 & 0.3929 & 0.8074
& 0.8671 & 0.8376 & 0.3661 & 0.8033
 \\
&  GAT
& 0.8204 & 0.7918 & 0.2827 & 0.6614
& 0.8646 & 0.8313 & 0.3720 & 0.7994
& 0.8536 & 0.8217 & 0.3557 & 0.7639

\\
\midrule
   \multirow{3}{*}{Four}  
&  GCN 
& 0.8333 & 0.7974 & 0.2964 & 0.6457
& 0.8722 & 0.8331 & 0.3799 & 0.7978
& 0.8667 & 0.8320 & 0.3560 & 0.7865

\\
&  GraphSAGE  
& 0.8373 & 0.8001 & 0.3073 & 0.6284
& 0.8768 & 0.8468 & 0.3966 & 0.8242
& 0.8706 & 0.8384 & 0.3768 & 0.7953
 \\
&  GAT
& 0.8258 & 0.7920 & 0.3072 & 0.6103
& 0.8657 & 0.8310 & 0.3851 & 0.7924
& 0.8621 & 0.8340 & 0.3652 & 0.7762

\\
\midrule
   \multirow{3}{*}{Eight}  
&  GCN 

& 0.8350 & 0.7994 & 0.3125 & 0.6272
& 0.8722 & 0.8380 & 0.3988 & 0.7958
& 0.8699 & 0.8327 & 0.3709 & 0.7807
\\
&  GraphSAGE  
& 0.8444 & 0.8107 & 0.3114 & 0.6306
& 0.8817 & 0.8523 & 0.4203 & 0.8209
& 0.8712 & 0.8427 & 0.3818 & 0.7840
 \\
&  GAT
& 0.8339 & 0.8007 & 0.2979 & 0.6118
& 0.8741 & 0.8425 & 0.4040 & 0.8156
& 0.8710 & 0.8415 & 0.3780 & 0.8183

\\
\bottomrule
   \end{tabular}
   \end{adjustbox}
   \caption{Results based on GNN type and number of demonstrations on using the CoSQL-S2T dataset.}
   \label{tab:gnn-type}
\end{table*}

\end{document}